\documentclass[10pt,journal,compsoc]{IEEEtran}
\usepackage{times}
\usepackage{epsfig}
\usepackage{graphicx, subfigure}
\usepackage{lscape}

\usepackage{graphicx}
\usepackage{amsmath}
\usepackage{amssymb}
\usepackage{amsthm}
\usepackage{cite}
\usepackage{url}
\usepackage{color}
\usepackage{verbatim}
\definecolor{Red}{rgb}{1, 0.2, 0.2}

\newcommand{\zzz}[1]{\textcolor{black}{#1}}
\newcommand{\aaa}[1]{\textcolor{black}{#1}}

\ifCLASSINFOpdf

\else

\fi

\begin{document}

\title{DeepPap: Deep Convolutional Networks for Cervical Cell Classification}

\author{Ling~Zhang,
        Le~Lu,~\IEEEmembership{Senior Member,~IEEE},
				Isabella~Nogues,
				Ronald~M.~Summers,
				\zzz{Shaoxiong~Liu,}
        and~Jianhua~Yao
\thanks{L. Zhang, L. Lu, R. Summers, and J. Yao are with the Imaging Biomarkers and Computer-Aided Diagnosis Laboratory and also with the Clinical Image Processing Service, Radiology and Imaging Sciences Department, National Institutes of Health Clinical Center, Bethesda, MD 20892 USA e-mail: (ling.zhang3@nih.gov; jyao@nih.gov).}
\thanks{I. Nogues is with the Imaging Biomarkers and Computer-Aided Diagnosis Laboratory, Radiology and Imaging Sciences Department, National Institutes of Health Clinical Center, Bethesda, MD 20892 USA.}
\thanks{\zzz{S. Liu is with the Department of Pathology, People’s Hospital of Nanshan District, Shenzhen 518052 China.}}
}


\maketitle

\begin{abstract}
Automation-assisted cervical screening via Pap smear \zzz{or liquid-based cytology (LBC)} is a highly effective cell imaging based cancer detection tool, where cells are partitioned into ''abnormal'' and ''normal'' categories. However, the success of most traditional classification methods relies on the presence of accurate cell segmentations. Despite sixty years of research in this field, accurate segmentation remains a challenge in the presence of cell clusters and pathologies. Moreover, previous classification methods are only built upon the extraction of \zzz{hand-crafted} features, such as morphology and texture. This paper addresses these limitations by proposing a method to directly classify cervical cells -- without prior segmentation -- based on \zzz{deep} features, using convolutional neural networks (ConvNets). First, the ConvNet is pre-trained on a natural image dataset. It is subsequently fine-tuned on a cervical cell dataset consisting of adaptively re-sampled image patches coarsely centered on the nuclei. In the testing phase, aggregation is used to average the prediction scores of a similar set of image patches. \zzz{The proposed method is evaluated on both Pap smear and LBC datasets. Results show that our method} outperforms previous algorithms in classification accuracy (98.3\%), area under the curve (AUC) (0.99) values, \zzz{and especially specificity (98.3\%),} when applied to the Herlev benchmark Pap smear dataset and evaluated using five-fold cross-validation. \zzz{Similar superior performances are also achieved on the HEMLBC (H\&E stained manual LBC) dataset.} Our method is promising for the development of automation-assisted reading systems in primary cervical screening.
\end{abstract}


\IEEEpeerreviewmaketitle

\section{Introduction}

\IEEEPARstart{C}{ervical} cytology (conventional Pap smear or liquid-based cytology) \cite{davey2006effect}, the most popular screening test for prevention and early detection of cervical cancer, has been widely used in developed countries and has significantly reduced its incidence and number of deaths \cite{saslow2012american}. However, population-wide screening is still unavailable in underdeveloped countries \cite{saslow2012american}, partly due to the complexity and tedious nature of manually screening abnormal cells from a cervical cytology specimen \cite{birdsong1996automated}. While automation-assisted reading techniques can boost efficiency, their current performance is not adequate for inclusion in primary cervical screening \cite{kitchener2011automation}. 

During the past few decades, extensive research has been devoted to the creation of  computer-assisted reading systems based on automated image analysis methods \cite{birdsong1996automated,bengtsson2014screening,zhang2014automation}. Such systems automatically select potentially abnormal cells in a given cervical cytology specimen, from which the cytoscreener/cytopathologist completes the classification. This task comprises three steps: cell (cytoplasm and nuclei) segmentation, feature extraction/selection, and cell classification.

Accurate cell segmentation is crucial to the success of a reading system. However, despite recent significant progress in this area \cite{gencctav2012unsupervised,plissiti2012overlapping,chen2014semi,zhang2014segmentation,chankong2014automatic,song2015accurate,lu2016evaluation,zhang2017graph,zhang2017combining}, the presence of cell clusters (which is even more problematic in Pap smear than in liquid-based cytology), as well as the large shape and appearance variations between abnormal and normal nuclei, remains a major obstacle to the accurate segmentation of individual cytoplasms and nuclei. On the Herlev benchmark dataset \cite{martin2003pap,jantzen2005pap}, the attained nucleus segmentation accuracy ranging between 0.85 \cite{chankong2014automatic} and \zzz{0.92 \cite{zhang2017combining}}. 
On an overlapping cervical cell dataset \cite{lu2016evaluation}, the cytoplasm segmentation accuracy ranges from 0.87 to 0.89 \cite{lu2016evaluation}. On the other hand, most cell classification studies assume that accurate segmentations of individual cytoplasms and nuclei are already available \cite{jantzen2005pap,marinakis2008particle,marinakis2009pap}. By optimizing features derived from the segmented cytoplasm and nucleus, high classification accuracies (e.g., 96.8\%) are achieved on the Herlev dataset, using 5-fold cross validation (CV)  \cite{marinakis2008particle,marinakis2009pap}. However, these high values would decrease, once the automated segmentation error (deriving mainly from the error-prone abnormal nucleus segmentation \cite{zhang2014segmentation,zhang2017graph}), were taken into account. 

Several strategies to remove this dependence on segmentation have been investigated. Classification based only on nucleus features (excluding cytoplasm features) is proposed \cite{plissiti2012importance,zhang2014automation,bora2017automated}. Comparable results on the Herlev dataset is obtained by using a non-linear dimensionality reduction algorithm and supervised learning \cite{plissiti2012importance}. Another idea is to classify image patches containing full cervical cells \cite{nanni2010local,guo2012discriminative,sokouti2014framework}. However, extraction of such patches still requires automated cell detection and segmentation. To avoid the pre-segmentation step, pixel-level classification method is designed to directly screen abnormal nuclei with no prior cytoplasm and nucleus segmentation \cite{zhang2004cervical}, but reports limited validation results. Alternatively, a technique which classifies the cropped blocks from cell images is proposed \cite{zhao2016automaticscreening}. However, arbitrary cropping could potentially separate a full cell into distinct patches.

Current cervical screening systems are hindered by limitations in the feature design and selection components. At present, extracted features fall under the following categories: handcrafted features describing morphology and chromatin characteristics \cite{jantzen2005pap,marinakis2008particle,marinakis2009pap,plissiti2012importance,chankong2014automatic} in accordance with ``The Bethesda System (TBS)" rules \cite{solomon2004bethesda}, engineered features representing texture distribution \cite{zhang2004cervical,nanni2010local,guo2012discriminative} according to previous computer-aided-diagnosis experiences, or both combined \cite{chen2014semi,zhang2014automation,zhao2016automaticscreening}. The resulting features are then organized, using a feature selection or dimensionality reduction algorithm, for classification. Handcrafted features are compromised by the current limited understanding of cervical cytology. Engineered features are obtained in an unsupervised manner, and thus often encode redundant information. The feature selection process potentially ignores significant clues and removes complementary information. Moreover, considering that the detection of some abnormal cervical cells is challenging even for human experts \cite{jantzen2005pap,solomon2004bethesda,desai2009role}, the \zzz{hand-crafted} features used in previous studies may not be able to represent complex discriminative information.  In fact, information describing cell abnormality may potentially lie in latent high\zzz{er} level features of cervical cell images, but this has not yet been investigated.

Representation learning refers to a set of methods designed to automatically learn and discover intricate discriminative information from raw data \cite{bengio2013representation}. Recently, representation learning has been popularized by deep learning methods \cite{lecun2015deep}. In particular, deep convolutional neural networks (ConvNets) \cite{lecun1989backpropagation} have achieved unprecedented results in the 2012 ImageNet Large Scale Visual Recognition Challenge, which consisted in classifying natural images in the ImageNet dataset into 1000 fine-grained categories \cite{krizhevsky2012imagenet}. They have also significantly improved performance in a variety of medical imaging applications \cite{greenspan2016guest,summers2016progress}, such as classification of lung diseases and lymph nodes in CT images \cite{shin2016deep,holger2016improving}, segmentation (pixel classification) of brain tissues \cite{moeskops2016automatic} in MRI, vessel segmentation \cite{liskowski2016segmenting} in fundus images, \zzz{and detecting cervical intraepithelial neoplasia (CIN, particularly CIN2+) at patient level based on Cervigram images \cite{xu2017multi} or multimodal data \cite{xu2016multimodal}}. Additionally, ConvNets have demonstrated superior performance in the classification of cell images, such as pleural cancer \cite{buyssens2012multiscale} and human epithelial-2 cell images \cite{gao2016hep}.

   \begin{figure}[!t]
   \begin{center}
   \begin{tabular}{c}
   \includegraphics[width=8.5cm]{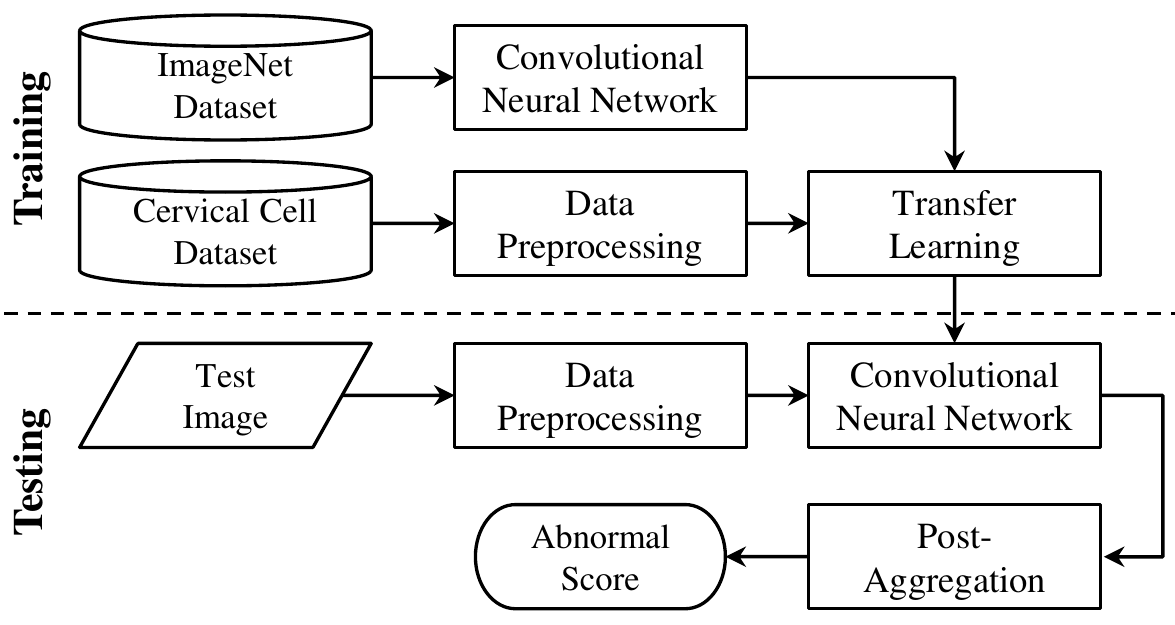}
   \end{tabular}
   \end{center}
   \caption[example] 
   { \label{figoverview} 
Overview of the proposed method using convolutional neural networks and transfer learning for classifying cervical cell images. 
}
   \end{figure} 

Large datasets are crucial to the high performance of Conv-Nets. However, there exists a very limited amount of labeled data for cervical cells, as high expertise is required for quality annotation. For instance, the Herlev benchmark dataset \cite{jantzen2005pap} only contains 917 cells (675 abnormal and 242 normal). Transfer learning \cite{yosinski2014transferable,bar2015chest,girshick2016region} is an effective method to overcome this problem. Since the features in the first few ConvNet layers are more generic, they can be appended to various sets of subsequent layers specific to different tasks \cite{yosinski2014transferable}. For instance, ConvNets trained on large-scale natural image datasets (e.g., ImageNet \cite{russakovsky2015imagenet}) can be transferred to various medical imaging datasets, such as CT \cite{shin2016deep}, ultrasound \cite{chen2015standard} and X-ray \cite{bar2015chest,carneiro2015unregistered} datasets, and can subsequently reduce overfitting on small datasets while boosting performance through fine-tuning.

In this paper, we apply ConvNets to the classification of cervical cells in cytology images. Our approach directly operates on raw RGB channels sampled from a set of square image patches coarsely centered on each nucleus. A ConvNet pre-trained on ImageNet is fine-tuned to discriminate between patches containing abnormal and normal cells \zzz{based on deep hierarchical features}. For an unseen cell, a set of image patches coarsely centered on the nucleus are classified by the fine-tuned ConvNet. Its classification results are then aggregated to generate the final cell category. Our approach is tested on \zzz{two cervical cell image datasets:} the Herlev dataset consisting of Pap smear images \cite{jantzen2005pap}; \zzz{the HEMLBC (H\&E stained manual liquid-based cytology) dataset being used to develop automation-assisted cervical screening system \cite{zhang2014automation}.} In our experiments (conducted using five-fold cross-validation(CV)), the fine-tuned ConvNet obtains classification \zzz{accuracies of 98.3\% on Herlev dataset and 98.6\% on HEMLBC dataset, surpass the previous best accuracies of 96.8\% and 94.3\% on the two datasets, respectively.} 

Our contributions are summarized as follows. 1) To the best of our knowledge, this is the first application of deep learning and transfer learning methods to cervical cell classification. 2) Unlike the previous methods, which rely on cytoplasm/nucleus segmentation and hand-crafted features, our method automatically extracts \zzz{deep hierarchical} features embedded in the cell image patch for classification, as long as a coarse nucleus center is provided. As a result, the classification does not suffer from any accuracy loss caused by inaccurate segmentation, and does not explicitly utilize prior medical knowledge of cervical cytology. 3) Our method generates the highest performances on \zzz{both} the Herlev Pap smear \zzz{and the HEMLBC datasets}, and has the potential to improve the performance of automation-assisted cervical cancer screening systems.

   \begin{figure*}[!t]
   \begin{tabular}{c}
   \includegraphics[width=17cm]{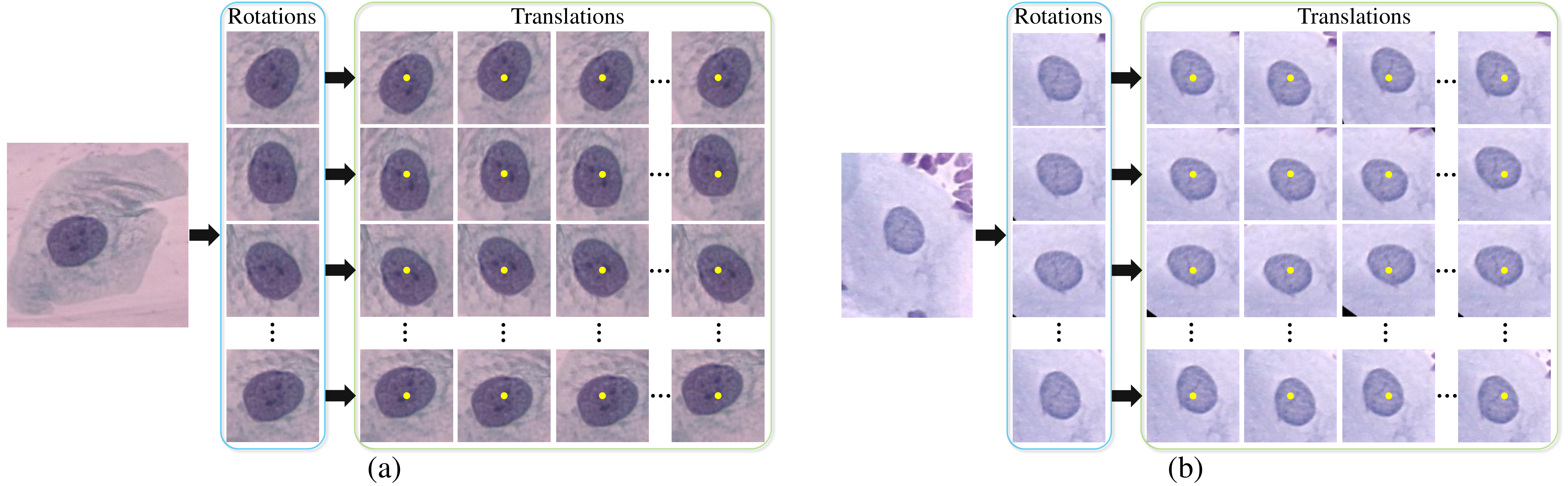}
   \end{tabular}
   \caption[example] 
   { \label{fig2} 
Two set of image patches are generated from (a) an abnormal and (b) a normal cervical cell image by rotations and translations. The centroids of translated image patches are shown as yellow, indicating the potentially inaccurate detection of nucleus centers.
}
   \end{figure*} 

\section{Methods}

The proposed method includes a training and a testing stage, as shown in Fig. \ref{figoverview}. In the training stage, a ConvNet is first pre-trained on the ImageNet dataset, and data preprocessing is applied on the \zzz{cervical cell} dataset. Next, transfer learning is applied, whereby the pre-trained network parameters are used to initialize a new ConvNet. This ConvNet is then fine-tuned on the preprocessed training samples. In the testing stage, the preprocessed testing images are fed into the fine-tuned ConvNet. The abnormality score is obtained by aggregating the ConvNet's output values. Further details are described below.

\subsection{Data Preprocessing}

\subsubsection{Patch extraction}
Unlike previous patch based cell classification methods \cite{nanni2010local,guo2012discriminative,sokouti2014framework,buyssens2012multiscale,gao2016hep}, our method does not directly operate on images containing full cells (like the images in the Herlev dataset), for practical reasons. In particular, obtaining an individual cell requires cell pre-segmentation (at least cytoplasm segmentation), which remains an unsolved, challenging problem \cite{lu2016evaluation}. As mentioned in the TBS rules \cite{solomon2004bethesda}, different cervical cytology abnormalities are associated with different nucleus abnormalities. Hence, nucleus features in themselves already include substantial discriminative information. We thus extract image patches of size $m \times m$ centered on the nucleus centroid. This strategy allows for embedding not only the nucleus scale/size information (an important discriminative feature between abnormal and normal cells), but also the contextual clues (e.g., the cytoplasm appearance) in the extracted patches. We acknowledge that automated methods for extracting a nucleus patch, e.g., Laplacian-of-Gaussian (LoG) \cite{lindeberg1998feature}, selective search \cite{uijlings2013selective}, or ConvNets \cite{xing2016automatic} exist. However, in this paper, we choose to focus on the classification task. We adopt a simple method of directly translating the centroid of the ground-truth nucleus mask to extract a set of image patches as described below. 

\subsubsection{Data Augmentation}
\label{augment}
Data augmentation improves the accuracy of ConvNets and reduces overfitting \zzz{\cite{krizhevsky2012imagenet}}. Since cervical cells are rotationally invariant, we perform $N_{r}$ rotations (with a step size of $\theta$ degrees) on each cell image, and thus increase our number of image samples. $N_{r}$ patches (one per rotated image) of size $m \times m$ centered at the rotated nucleus centers are extracted as the training samples, as shown in the middle (blue) panel in Fig.\ \ref{fig2}. \zzz{Note that rotating a cell image may slightly degrade its high frequency contents (could be considered as a lower imaging quality), but should not change its abnormality/normality for most cells.} \zzz{Actually the augmentation} step \zzz{based on image rotation} is crucial to the success of the ConvNet \cite{shin2016deep}, \zzz{and has been demonstrated to be important for improving accuracy of ConvNet-based cell image classification problem \cite{gao2016hep}}, given the limited number of images in the Herlev \zzz{and HEMLBC} dataset. Zero padding is used to void regions that lie outside of the image boundary.

Considering that the detected nucleus center may be inaccurate in practice, we randomly translate (by up to $d$ pixels) each nucleus centroid $N_{t}$ times to obtain $N_{t}$ points as the coarse nucleus centers. Accordingly, $N_{t}$ patches of size $m \times m$ centered at these locations are extracted as training samples, as shown in the right (green) panel in Fig.\ \ref{fig2}. These patches not only simulate inaccurate nucleus center detection, but also increase the amount of training samples for ConvNets. Other data augmentation approaches such as scale and color transformations are not used, as both the size and intensity of the nucleus are essential features to distinguish abnormal cervical cells from normal ones.

There are $\sim 3$ times more abnormal cell images than normal cell images in the Herlev dataset. Classifiers tend to exhibit bias towards the majority class (abnormal cells). Although achieving a high sensitivity rate (correct classification of abnormal cells) is ideal from a medical point of view \cite{jantzen2005pap}, a high false positive rate (mis-classification of normal cells as abnormal) is not desirable from a practical standpoint \cite{zhang2014automation}. A common solution to this dilemma is to balance the proportions of positive and negative training samples \cite{he2009learning}. Doing so also improves the accuracy and convergence rate of ConvNets in training \cite{krizhevsky2012imagenet,holger2016improving}. Hence, we create a balanced training set by sampling a higher proportion of normal than abnormal patches.

\subsection{Convolutional Neural Networks}
\begin{figure*}[!t]
   \begin{tabular}{c}
   \includegraphics[width=17cm]{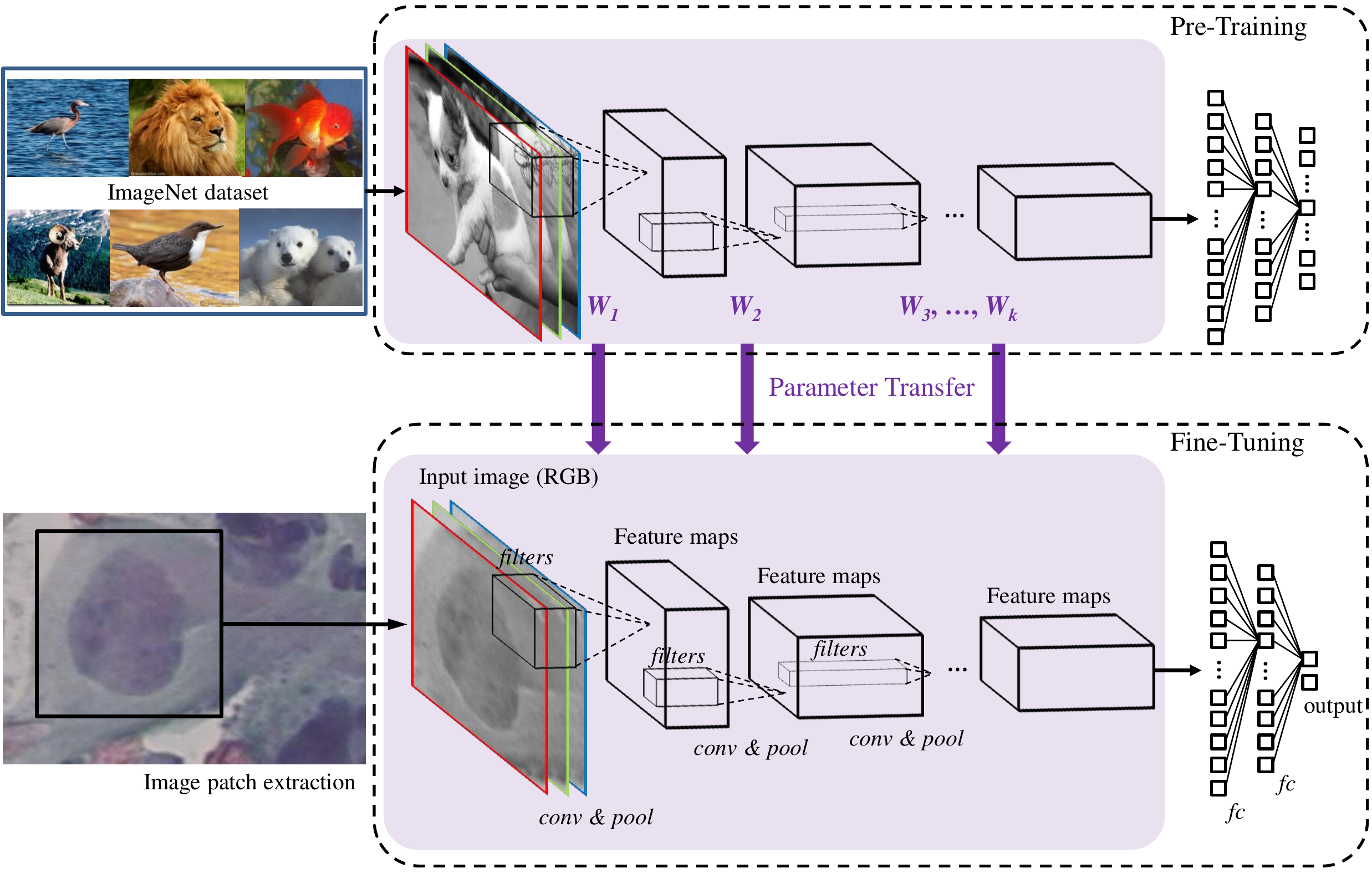}
   \end{tabular}
   \caption[example] 
   { \label{fig3} 
A ConvNet is fine-tuned on the cervical cell dataset with parameters transferred from another ConvNet pre-trained on the ImageNet dataset. In this example, the parameters ($W_{1}$, $W_{2}$, ... $W_{k}$) in the purple region in the pre-trained model (upper panel) are transferred to the same locations in another model (lower panel) for fine-tuning on cervical cell dataset.
}
   \end{figure*}

A convolutional neural network (ConvNet) \cite{lecun1989backpropagation,krizhevsky2012imagenet} is a deep learning model comprising multiple consecutive stages, namely \textit{convolutional} ($conv$), \textit{non-linearity} and \textit{pooling} ($pool$) layers, followed by more $conv$ and \textit{fully connected} ($fc$) layers. The input of the ConvNet is the raw pixel intensity image (in our case, the image obtained by subtracting the mean image over the training set from the original image \cite{krizhevsky2012imagenet}). The output layer is composed of several neurons each corresponding to one class. The weights ($W$) in the ConvNet are optimized by minimizing the classification error on the training set using the backpropagation algorithm \cite{lecun2012efficient}. Fig. \ref{fig3} shows two ConvNets. The upper network is trained on the ImageNet dataset, and the lower network is trained on the cervical cell dataset.

\subsubsection{Convolutional layer}
The $conv$ layer takes local rectangular patches across (with offset by \textit{stride} and with/without spatial preservation by \textit{padding}) the input image (for the first layer) or feature maps (for the subsequent layers) as input, on which 2D convolution with a filter is performed. The sum $x$ of the resulting convolutions is fed into a non-linearity function, specifically a rectified linear unit (ReLU) $f(x) = max(0,x)$ \cite{krizhevsky2012imagenet}, in order to increased the speed of training. In a given layer, the same filter is shared in a feature map, while different filters are used for different feature maps. This property of filter sharing in $conv$ layer allows for detecting the same pattern in different locations of the feature map.

\subsubsection{Pooling layer}
The pooling operation down-samples the feature map by summarizing feature responses in each non-overlapping local patch, often by computing the maximum activations (max-pooling). This yields features invariant to minor translations in the data. 

\subsubsection{Fully connected layer}
$conv$ and $pool$ generate feature maps of smaller dimensions than the input image, which are then passed through several $fc$ layers. The first few $fc$ layers fuse these feature maps into a feature vector. The last $fc$ layer contains two neurons which compute the classification probability for each class using softmax regression. To reduce overfitting, ``dropout" \cite{hinton2012improving} is used to constrain the fully-connected layers.

\subsubsection{Network training}
The weights $W$ in ConvNets are initialized with values from the Gaussian distribution. During training, these weights are iteratively updated with the gradients of the loss function, computed via stochastic gradient descent (SGD) over a mini-batch (size of 256) of training samples. The initial learning rate is decreased after certain epochs. As in Ref. \cite{krizhevsky2012imagenet}, momentum and weight decay are used to speed up the learning and reduce overfitting. The training process is terminated after a pre-determined number of epochs. The model with the lowest validation loss value is selected as the final network.

\subsection{Transfer Learning}
Transfer learning refers to the fine-tuning of deep learning models that are pre-trained on other large-scale image datasets. In this study, the first few $conv$ and $pool$ layers of a ConvNet pre-trained on the ImageNet classification dataset (ILSVRC2012) (purple region in the upper part in Fig. \ref{fig3}) are used as the base of our network, on top of which several task-specific $fc$ layers with random initialized weights are attached. In order to facilitate the transfer of features, the same network layers ($conv$ and $pool$) with the BVLC CaffeNet \cite{jia2014caffe} are transferred to the same locations in our model (purple regions in Fig. \ref{fig3}). Like our network, the CaffeNet also takes RGB channels as input. All of these layers are jointly trained (fine-tuned) on our cervical cell dataset, for which a learning rate 10 times smaller than the original CaffeNet value is used to fine-tune the transferred layers, and the original learning rate is used to train the $fc$ layers from scratch.

\subsection{\zzz{Testing}}
\label{classify}
To classify an unseen image, we combine random-view aggregation \cite{holger2016improving} and multiple crop testing \cite{krizhevsky2012imagenet} to produce the final prediction score. In particular, our data augmentation method generates $N_{v}$ image patches (rotations and translations about the nucleus centroid). From each of these patches, $N_{c}$ sub-patches are cropped (including its corner, center and mirrored patches). Hence, for each test cervical cell image, $N_{v} \times N_{c}$ sub-patches are fed into the ConvNet. The final prediction score is obtained by averaging the scores of these $N_{v} \times N_{c}$ predictions.

\section{Experimental Methods}

\subsection{Data set}
\zzz{The cell data used to train and test the ConvNets comes from two datasets with two types of cervical cytology images, which were acquired by different slide preparation, staining methods, and imaging conditions.}

\subsubsection{Herlev Dataset}
\begin{figure*}[!t]
   \begin{center}
   \begin{tabular}{c}
   \includegraphics[width=15cm]{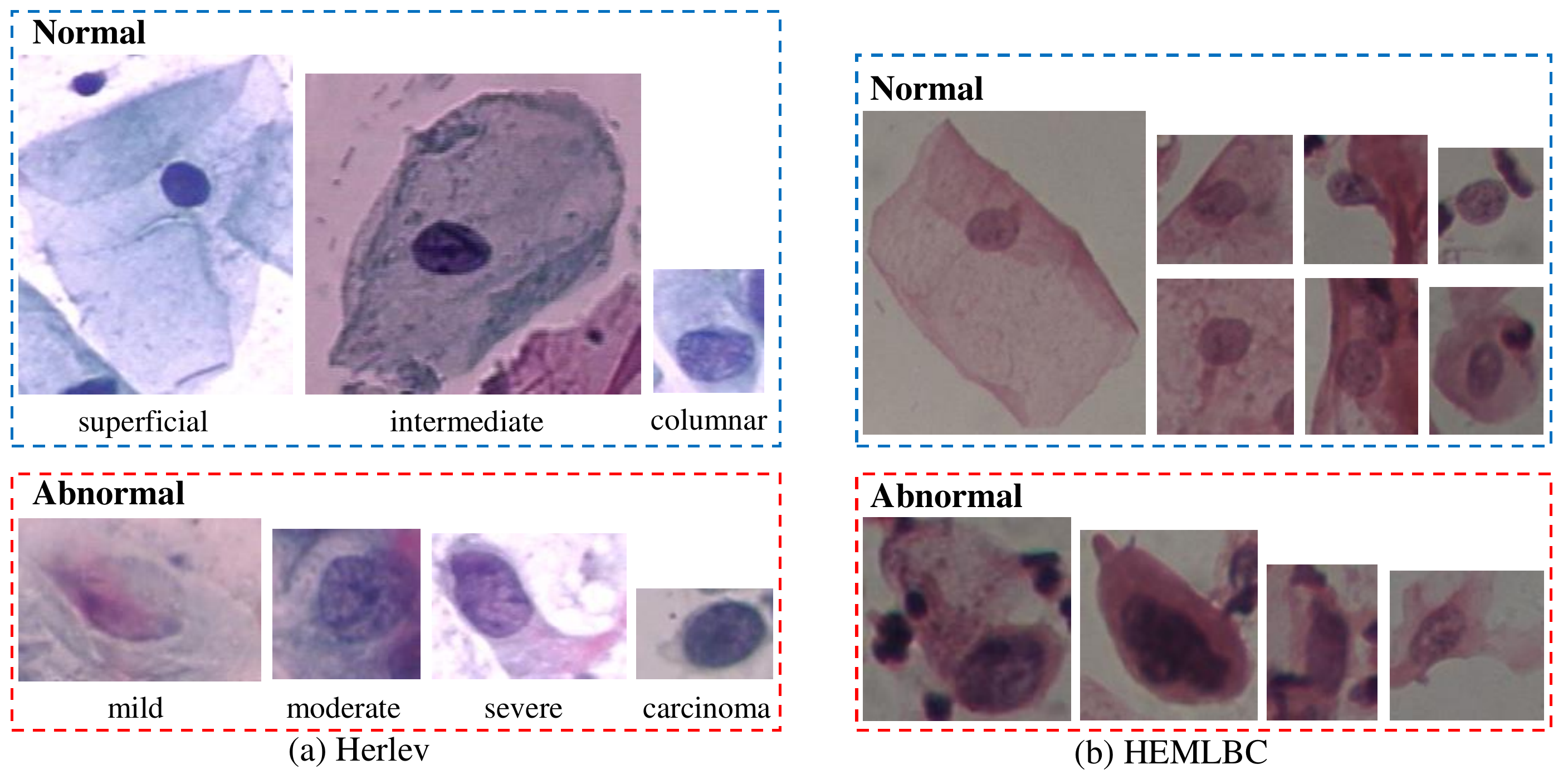}
   \end{tabular}
   \end{center}
   \caption[example] 
   { \label{fig1} 
Example images of normal and abnormal cervical cells from the \zzz{(a)} Herlev \zzz{and (b) HEMLBC} dataset. All these examples keep their originally relative scales for better illustrating the \zzz{different characteristics (mainly nucleus size)} between normal and abnormal cells. 
}
   \end{figure*}

The \zzz{first one} is from a publicly available dataset \zzz{(http://mde-lab.aegean.gr/downloads)} collected at the Herlev University Hospital by a digital camera and microscope \cite{jantzen2005pap}. \zzz{The image resolution is 0.201 $\mu m$ per pixel \cite{martin2003pap}}. The specimens are prepared via conventional Pap smear \zzz{and Pap staining}. The Herlev dataset consists of 917 images -- each containing one cervical cell -- with ground truth segmentation and classification. There are a total of seven different classes -- diagnosed by two cyto-technicians and a doctor, in order to maximize certainty of the diagnosis. These seven classes belong to two categories: class 1-3 are normal, and class 4-7 are abnormal, as shown in Table \ref{herlevdataset}. Examples of some cells are provided in Fig. \ref{fig1}\zzz{(a)}. \zzz{As can be seen, most abnormal cells have larger nucleus size than normal cells. However, the normal columnar nucleus may have similar size (also maybe similar chromatin distribution) as severe and/or carcinoma nuclei, which makes the classification challenging.}

\begin{table}[!t]
\caption{The 917 cells (242 normal and 675 abnormal) from Herlev dataset.
}
\label{herlevdataset}
\begin{tabular}{|p{1cm}|p{0.5cm}|p{5.4cm}|p{0.5cm}|}
\hline
Category & Class & Cell type & Num.\\
\hline
Normal & 1 & Superficial squamous epithelial & 74\\
\hline
Normal & 2 & Intermediate squamous epithelial & 70\\
\hline
Normal & 3 & Columnar epithelial & 98\\
\hline
Abnormal & 4 & Mild squamous non-keratinizing dysplasia & 182\\
\hline
Abnormal & 5 & Moderate squamous non-keratinizing dysplasia & 146\\
\hline
Abnormal & 6 & Severe squamous non-keratinizing dysplasia & 197\\
\hline
Abnormal & 7 & Squamous cell carcinoma in situ intermediate & 150\\
\hline
\end{tabular}
\end{table}

For each abnormal cell image in the Herlev dataset, $N_{r} = 10$ rotations ($\theta$ = 36$^{\circ}$) and $N_{t} = 10$ translations (up to 10 pixels) are performed. For each normal cell, we use $N_{r} = 20$ ($\theta$ = 18$^{\circ}$) and $N_{t} = 14$, resulting in 100 and 280 image patches for each abnormal and normal cell image, respectively. This yields a relatively balanced data distribution. \zzz{Note that such different steps of rotation/translation for abnormal and normal cells are only for training not testing set.} The image patch size is set to $m = 128$ pixels to cover some cytoplasm region for most cells, and to contain most of the full nucleus region for the largest one. These image patches are then up-sampled to a size of 256 $\times$ 256 $\times$ 3 pixels via bi-linear interpolation, \zzz{in order to facilitate the transfer of pre-trained ConvNet model} \cite{shin2016deep}.

\zzz{
\subsubsection{HEMLBC Dataset}
The second one is from our own dataset collected at the People's Hospital of Nanshan District by using our previously developed autofocusing system (Olympus BX41 microscope with 20$\times$ objective, Jenoptik ProgRes CF Color 1.4 Megapixel Camera, and MS300 motorized stage) \cite{zhang2014automation}. Each pixel has a size of 0.208 $\mu m^{2}$. The specimens are prepared by manual liquid-based cytology with H\&E staining. The dataset used in this paper is a subset used to train the abnormal/normal nucleus classifier for our automation-assisted cervical screening system \cite{zhang2014automation}. There are totally 989 abnormal cells from 8 biopsy-confirmed CIN slides and 1381 normal cells from another 8 NILM (negative for intraepithelial lesion and malignancy) slide available. To create a balanced data distribution, 989 normal cells are randomly selected. The abnormal cells are diagnosed by two experienced pathologists. Most of them are segmented by an automated algorithm \cite{zhang2014automation} and the ill-segmented ones are manually segmented by a pathologist. The normal cells are formed by two subsets: the first subset is collected by a pathologist with automated segmentation; the second subset is some false positive cells (e.g., cells with large nuclei, atrophic cells, etc.) collected during bootstrap process from validation images with manual segmentation for the ill-segmented ones by an engineer. More details are described in \cite{zhang2014automation}. Examples of some cells are shown in Fig. \ref{fig1}(b).
}

\zzz{
For both abnormal and normal cell image in the HEMLBC dataset, $N_{r} = 10$ rotations ($\theta$ = 36$^{\circ}$) and $N_{t} = 10$ translations (up to 10 pixels) are performed, resulting in 100 image patches for each cell image. The image patch size is also set to $m = 128$ pixels and then up-sampled to a size of 256 $\times$ 256 $\times$ 3 pixels as in Herlev dataset.
}

\subsection{Network Architectures and Implementation}
Fig. \ref{fig3} illustrates our network architecture. The base ConvNet (denote as ConvNet-B) is pre-trained on the ImageNet classification dataset. ConvNet-B contains five $conv$ layers ($conv1 - conv5$), three $pool$ layers ($pool1$, $pool2$, $pool5$), and three $fc$ layers ($fc6 - fc8$). Layers from $conv1$ to $pool5$ are transferred to the same locations in our model (denote as ConvNet-T). \zzz{In other word, the first 5 weight layers ($conv1$ to $pool5$) of ConvNet-T are copied from the pre-trained ConvNet-B, and $fc6 - fc8$ layers of ConvNet-T are initialized with random Gaussian distributions. The detailed configurations of our ConvNet-T are listed in Table. \ref{parameter}. \zzz{Local response normalization is used for $conv1$ and $conv2$ layers using the same setting as \cite{krizhevsky2012imagenet}, and all hidden layers are equipped with the ReLU activation function. }Note that the ConvNet-B and ConvNet-T share the same network structure from $conv1$ to $pool5$, but the number of neurons of the three $fc$ layers in ConvNet-B and ConvNet-T are 4096-4096-1000 and 1024-256-2, respectively. The 1024 and 256 are set based on our empirical evaluation, and 2 is to accommodate the new object categories in our 2-class (abnormal/normal) classification problem. Actually, setting the number of neurons of $fc6$ and $fc7$ layers in the range of 1024$\sim$256 will result in similar accuracy, while more number of neurons (e.g., 4096-4096) tend to have slightly lower accuracy (1\%-2\% lower) on our data, which is more compact and specific compared to ImageNet.} ConvNet-T is run on the Caffe platform \cite{jia2014caffe}, using a  Nvidia GeForce GTX TITAN Z GPU with \zzz{12} GB of memory. 

\begin{table*}[!t]
\centering
\caption{Con\zzz{v}Net-T architectures for cervical cell classification.
}
\label{parameter}
\begin{tabular}{|p{1.2cm}|p{0.7cm}|p{0.7cm}|p{0.7cm}|p{0.7cm}|p{0.7cm}|p{0.7cm}|p{0.7cm}|p{0.7cm}|p{0.7cm}|p{0.7cm}|p{0.7cm}|p{0.7cm}|}
\hline
 & Input & $conv1$ & $pool1$ & $conv2$ & $pool2$ & $conv3$ & $conv4$ & $conv5$ & $pool5$ & $fc6$ & $fc7$ & $fc8$\\
\hline
Filter size & - & 11$\times$11 & 3$\times$3 & 5$\times$5 & 3$\times$3 & 3$\times$3 & 3$\times$3 & 3$\times$3 & 3$\times$3 & - & - & \\
\hline
Channels & 3 & 96 & 96 & 256 & 256 & 384 & 384 & 256 & 256 & 1024 & 256 & 2\\
\hline
Stride & - & 4 & 2 & 1 & 2 & 1 & 1 & 1 & 2 & - & - & -\\
\hline
Padding & - & - & - & 2 & - & 1 & 1 & 1 & - & - & - & -\\
\hline
\end{tabular}
\end{table*}

\subsection{Training and Testing Protocols}
From each 256 $\times$ 256 training image patch or its mirrored version, a 227 $\times$ 227 sub-patch is randomly cropped, from which \zzz{the mean image over the training set is then subtracted.} 
Stochastic Gradient Descent (SGD), with a mini-batch size of 256, is used to train the ConvNet-T for 30 epochs. The learning rates of layers $conv1 - pool5$ and layers $fc6 - fc8$ start from 0.001 and 0.01, respectively, and are decreased by a factor of 10 at every tenth epoch. Weight decay and momentum are set to 0.0005 and 0.9. A dropout ratio of 0.5 is used to constrain the $fc6$ and $fc7$ layers.

In testing, we obtain the final score by averaging the scores of the $fc8$ output on 1000 patches ($N_{v}$ = 100  augmentations each with $N_{c}$ = 10 sub-crops).

\subsection{Evaluation Methods}
We evaluate the cervical cell classification using five-fold CV \zzz{on both Herlev and HEMLBC datasets}, to facilitate comparison with most previously reported results. In each of the 5 iterations of the ConvNet, 4 of 5 folds are used as training data, and the remaining one as validation. 
\aaa{It's worth mentioning that data augmentation is after the training/validation splitting of cell population.}
We obtain the model's final performance values by averaging results from the 5 validation sets. The performance evaluation metrics include sensitivity ($Sens$), specificity ($Spec$), accuracy ($Acc$), harmonic mean ($H$-$mean$), \zzz{$F$-$measure$,} and area under the ROC curve ($AUC$),
where $Sens$ measures the proportion of correctly identified abnormal cells, and $Spec$ the proportion of correctly identified normal cells; $Acc$ is the global percentage of correctly classified cells; $H$-$mean$ = $2 \times \frac{Sens\times Spec}{Sens+Spec}$, used in \cite{plissiti2012importance}, takes into account the imbalanced data distribution\zzz{; $F$-$measure$, the harmonic mean of precision and recall, is used in \cite{bora2017automated}.} The ROC curve is computed by varying thresholds on the final classification scores (each final score is the average score of 1000 predictions).
 \zzz{To test the robustness of our method against localization error of nucleus center, we randomly translate the ground truth centers of the test cells up to 5 or 10 pixels in both $x$ and $y$ directions, and the resulting performances on Herlev dataset are reported.} In addition, the numbers of correct classification (normal vs. abnormal) \zzz{and the distribution (shown by box plots) of the predicted abnormal scores of all cells} for each of the seven cell classes (listed in Table \ref{herlevdataset}) are reported.
\aaa{Finally, we further consider the 7-class classification problem by simply modifying the number of neurons in the last $fc$ layer from 2 to 7, and report the overall error (OE)\% as in \cite{jantzen2005pap,marinakis2009pap}.}

\section{Results}
\begin{figure}[!t]
   \includegraphics[width=8.8cm]{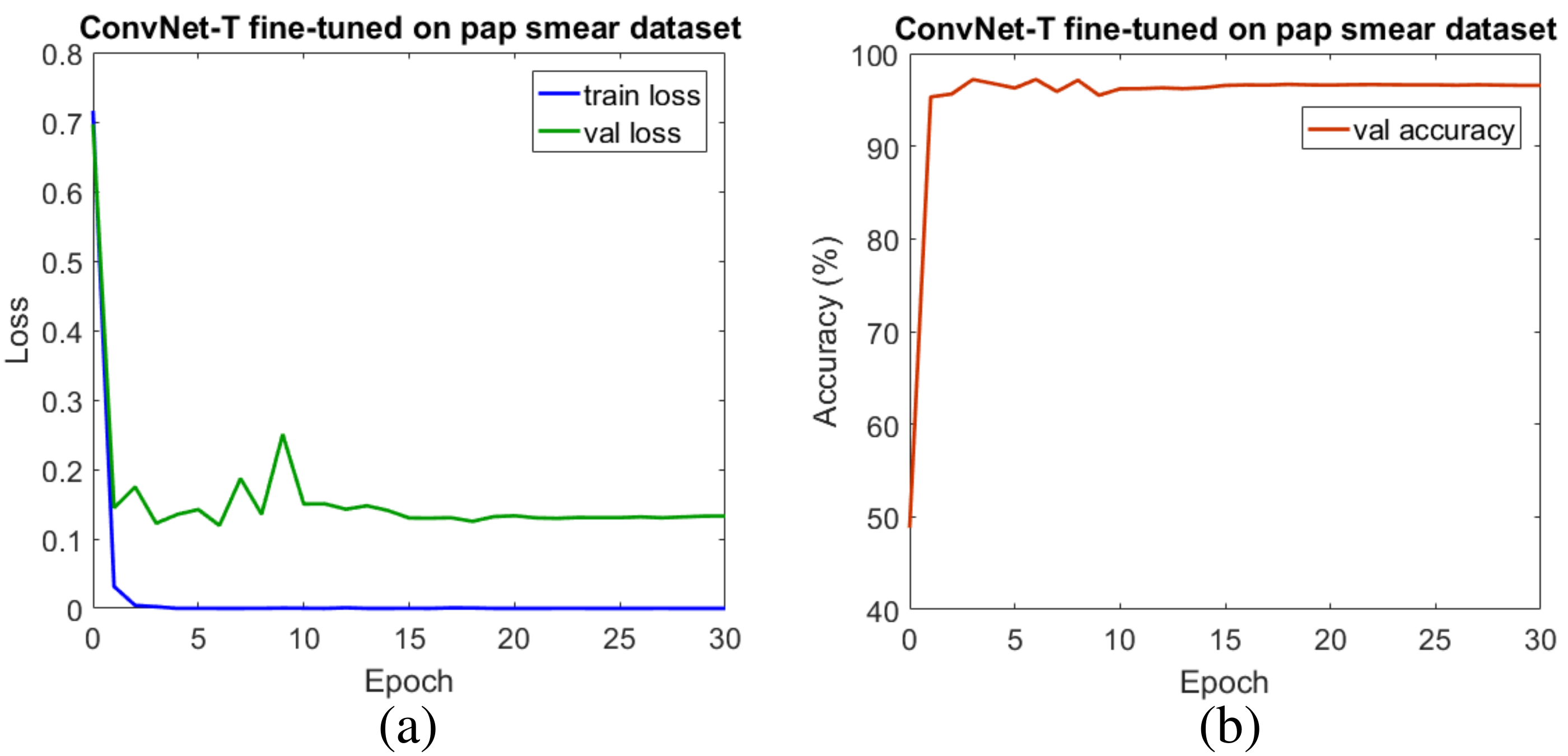}
   \caption[example] 
   { \label{figlossacc} 
\zzz{(a)} Training and validation loss, and \zzz{(b)} validation accuracy versus number of training epoch\zzz{s}.
}
   \end{figure}

\begin{figure}[!t]
   \begin{center}
   \begin{tabular}{c}
   \includegraphics[width=8.5cm]{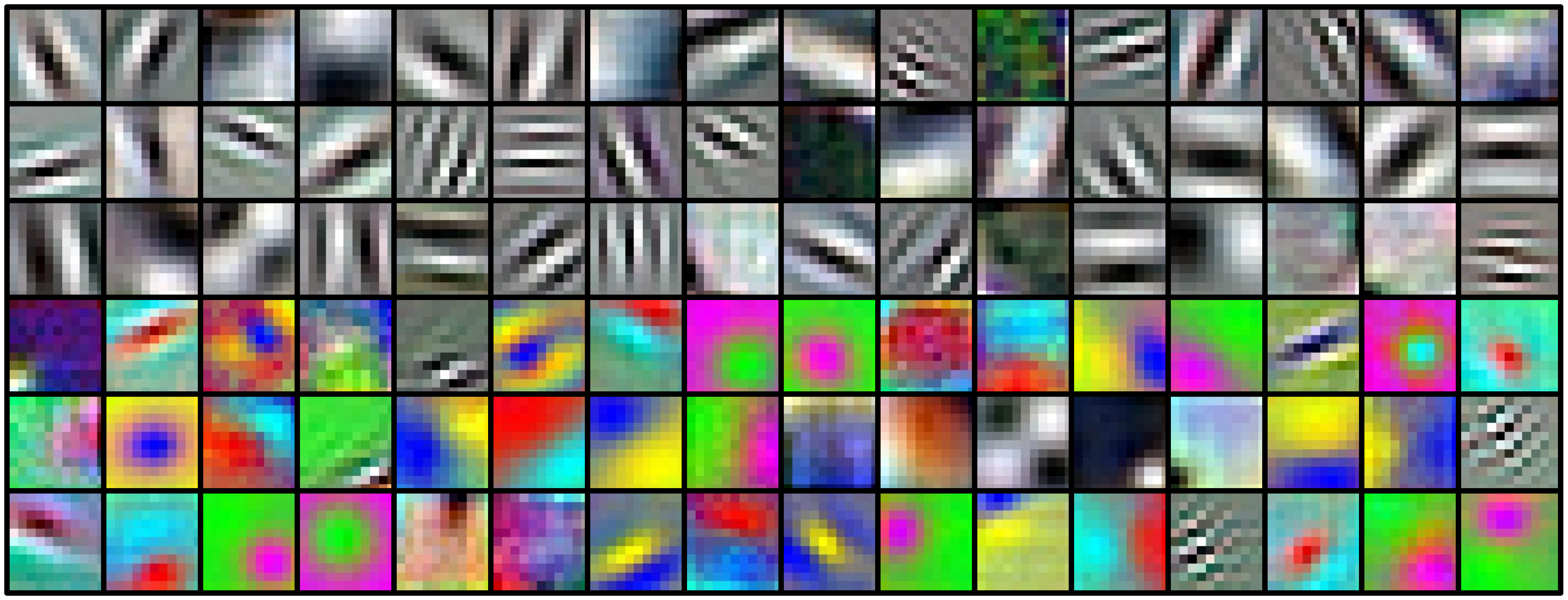}
   \end{tabular}
   \end{center}
   \caption[example] 
   { \label{figconv} 
Visualization of the 96 filters with size of 11 $\times$ 11 $\times$ 3 in the first convolutional layer of ConvNet-T fine-tuned on Pap smear dataset.
}
   \end{figure}

\begin{figure*}[!htbp]
   \begin{center}
   \begin{tabular}{c}
   \includegraphics[width=17.5cm]{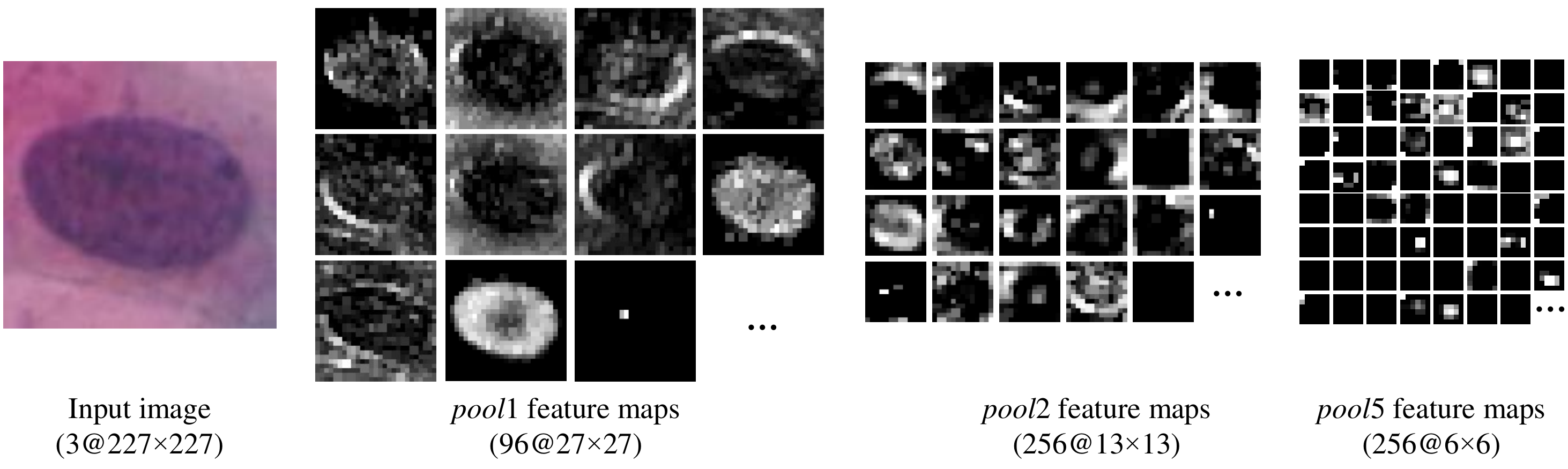}
   \end{tabular}
   \end{center}
   \caption[example] 
   { \label{figfeaturemaps} 
Visualization of the activations (feature maps) of three pooling layers, $pool1$, $pool2$, and $pool5$ for an input cervical cell image.
}
   \end{figure*}

\zzz{\subsection{ConvNet Learning Results}}
Fig. \ref{figlossacc} illustrates a fine-tuning process of ConvNet-T during 30 training epochs \zzz{on the Herlev dataset}. As shown in the figure, after 6 epochs, the validation loss reaches its minimum value (0.119), with a corresponding validation accuracy of 0.972. Fig. \ref{figconv} shows the learned filters of the first convolutional layer of ConvNet-T trained on the Herlev Pap smear dataset. These automatically learned filters mainly consist of gradients of different frequencies and orientations and blobs of color, which are necessary for the cervical cell classification task. Along with these learned filters, the activations (feature maps) of an example cell at different pooling layers ($pool1$, $pool2$, and $pool5$) are provided in Fig. \ref{figfeaturemaps}. One can observe that the pooling operation summarizes the input cell image or previous feature maps by highlighting the activated spatial locations, and that the features become increasingly abstract in deeper layers of the ConvNet. 

\begin{figure}[!t]
   \begin{center}
   \begin{tabular}{c}
   \includegraphics[width=8.5cm]{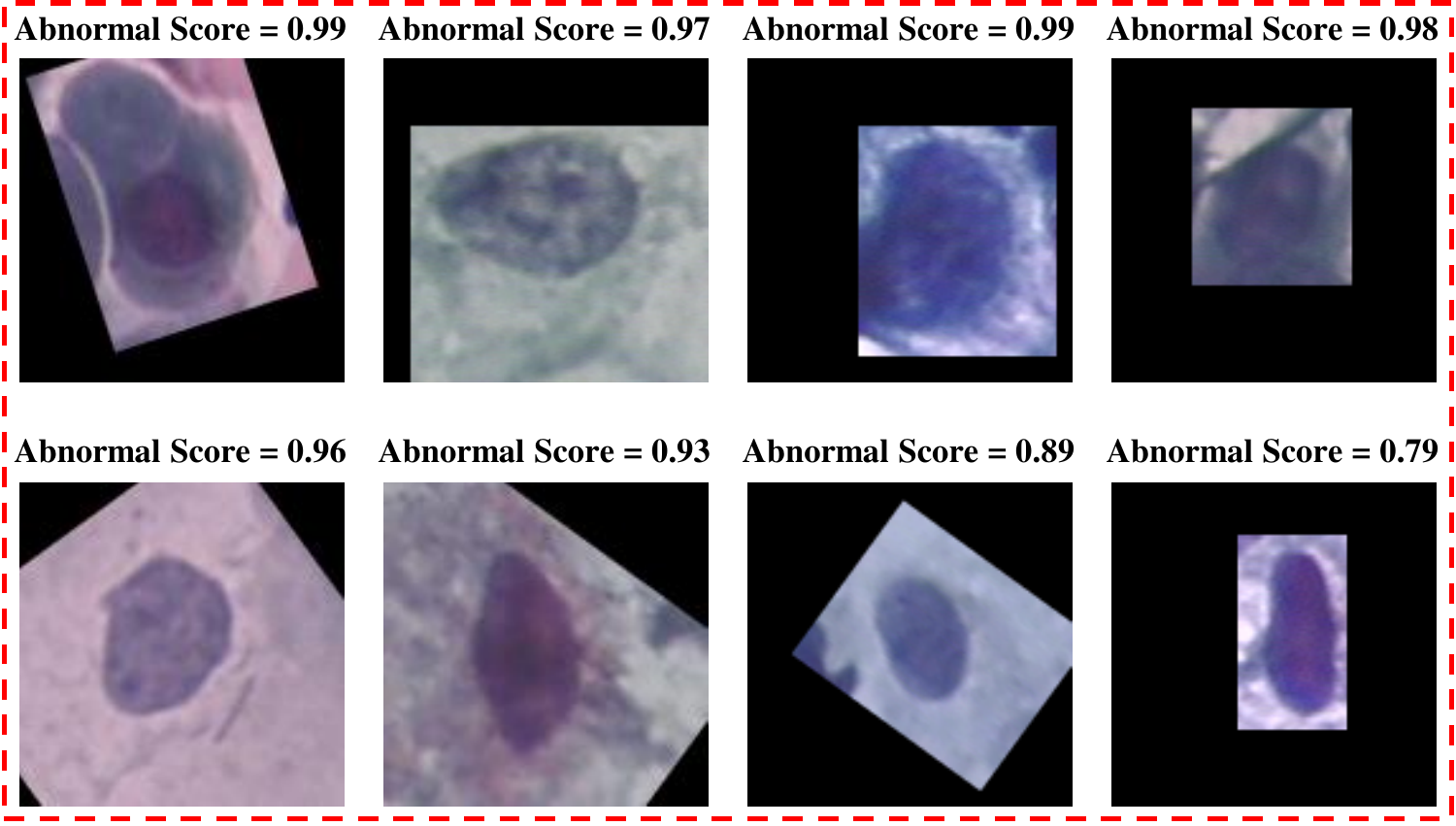}
   \end{tabular}
   \end{center}
   \caption[example] 
   { \label{fig4} 
Examples of correctly classified abnormal cervical cells \zzz{from Herlev dataset}. Column 1 to column 4 are mild dysplasia, moderate dysplasia, severe dysplasia and carcinoma, respectively. Score = 1 corresponds to a 100\% probability of representing an abnormal cell.
}
   \end{figure}

\begin{figure}[!t]
   \begin{center}
   \begin{tabular}{c}
   \includegraphics[width=6.5cm]{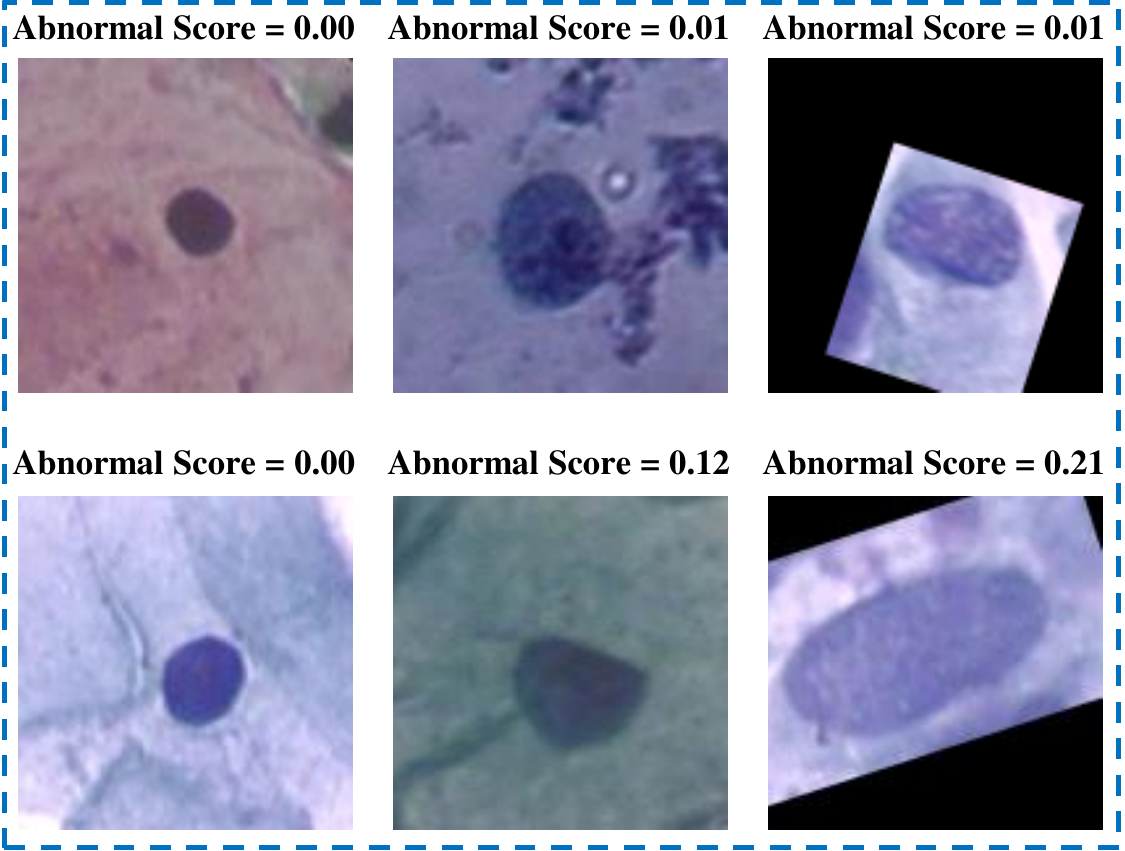}
   \end{tabular}
   \end{center}
   \caption[example] 
   { \label{fig5} 
Examples of correctly classified normal cervical cells \zzz{from Herlev dataset}. Column 1 to column 3 are superficial, intermediate and columnar epithelial, respectively. Score = 1 corresponds to a 100\% probability of representing an abnormal cell.
}
   \end{figure}

\begin{figure}[!t]
   \begin{tabular}{c}
   \includegraphics[width=8.8cm]{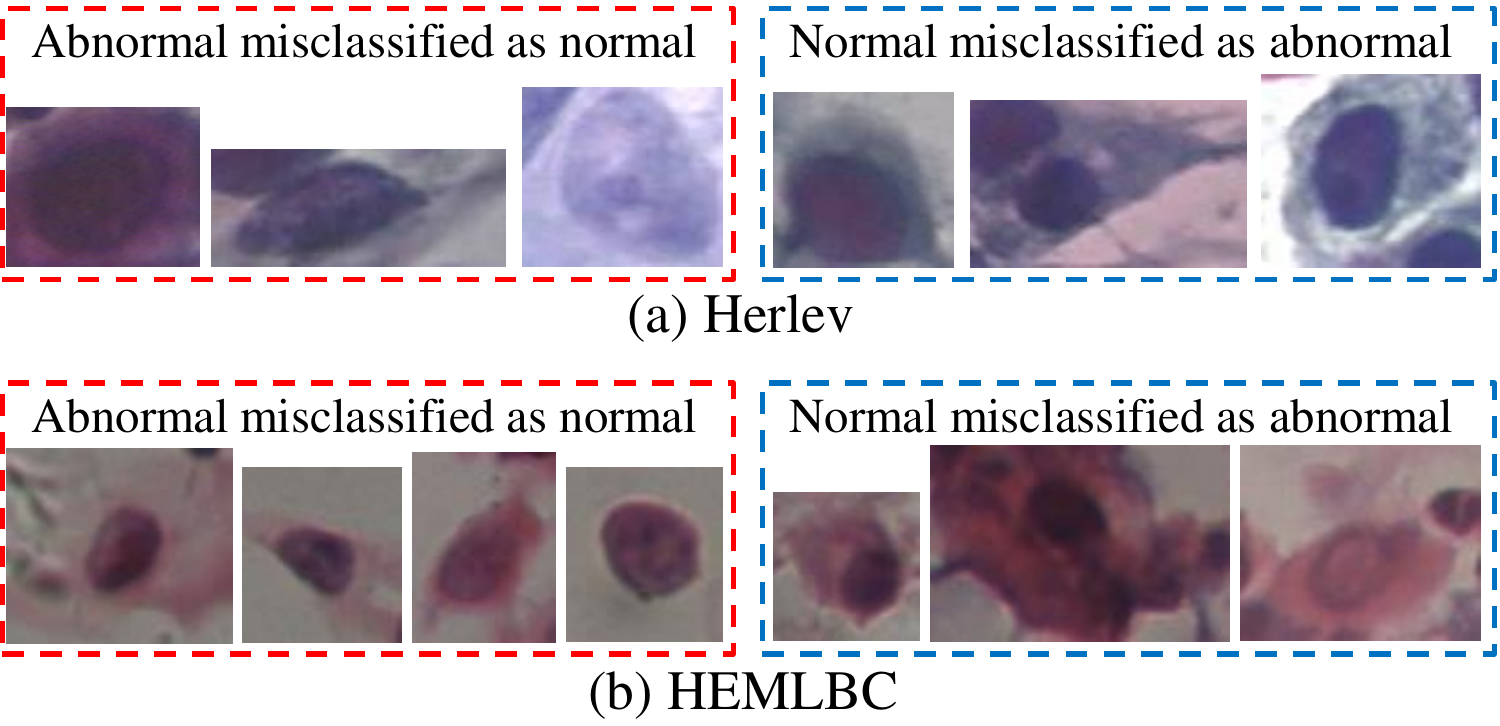}
   \end{tabular}
   \caption[example] 
   { \label{fig6} 
Examples of misclassified cervical cells \zzz{from (a) Herlev dataset; (b) HEMLBC dataset.} All images are shown at their original scales.
}
   \end{figure}

\zzz{\subsection{Qualitative Results}}
Fig. \ref{fig4} and Fig. \ref{fig5} contain examples of correctly classified abnormal and normal cell patches from the validation \zzz{Herlev dataset}, respectively. Fig. \ref{fig6} provides examples of misclassified cervical cells \zzz{from both Herlev and HEMLBC datasets}, including both false negatives and false positives. The first two false negatives are instances of carcinoma, and the third one is an example of severe dysplasia. All false positives are columnar epithelial cells. 

\zzz{\subsection{Quantitative Results on Herlev Dataset}}
Table \ref{performance} shows the classification performance ($Sens$, $Spec$, $Acc$, $H$-$mean$, \zzz{$F$-$measure$,} and $AUC$) of our method in comparison with previous methods \cite{chankong2014automatic,jantzen2005pap,marinakis2008particle,marinakis2009pap,plissiti2012importance,nanni2010local,guo2012discriminative} on the Herlev dataset. 
The mean values of $Sens$, $Spec$, $Acc$, $H$-$mean$, \zzz{$F$-$measure$,} and $AUC$ from our method (ConvNet-T) are 98.2\%, 98.3\%, 98.3\%, 98.3\%, \zzz{98.8\%,} and 0.99\zzz{8}, respectively. We thus outperform previous methods in all metrics but $Sens$, which is slightly below others. Among these metrics, our $Spec$ (98.3\%) substantially surpasses the previous highest result (92.2\%). \zzz{Also note that certain degree of localization error (up to 10 pixels) of nucleus center only results in a small reduction of performances of our method (e.g., $Acc$ from 98.3\% to 97.8\%).}

\begin{figure}[!t]
   \begin{center}
   \includegraphics[width=8.8cm]{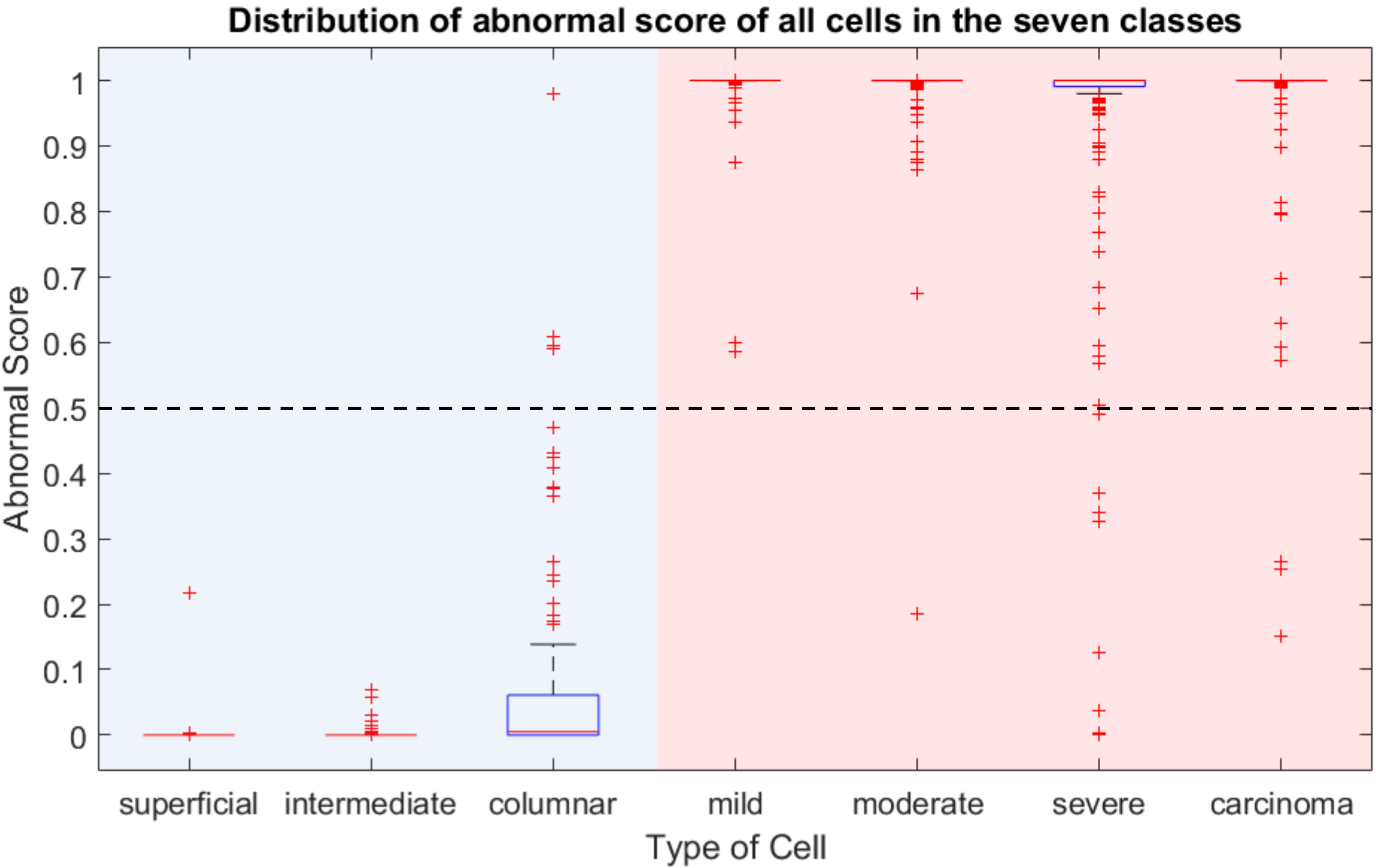}
   \end{center}
   \caption[example] 
   { \label{figdistrib} 
\zzz{ Box plots shown the distribution of abnormal scores of all cells in the seven classes from Herlev dataset. Blue and red backgrounds indicate normal and abnormal cell categories, respectively. Black dash line is the score threshold 0.5.}
}
   \end{figure}

Table \ref{confusionmatrix} provides the numbers (and corresponding percentages) of correct classification for each of the seven cell classes. Our method shows perfect performance on two types of normal cell (superficial and intermediate squamous epithelial), as well as one type of abnormal cell (mild dysplasia). While the performances are relatively lower for columnar epithelial and severe squamous non-keratinizing dysplasia (both are 95.9\%). \zzz{The distribution of the abnormal scores of all cells for the seven cell classes are shown as box plots in Fig. \ref{figdistrib}. The proposed method returns abnormality-scores close to 0 or 1 for most normal and abnormal cells, respectively. The few misclassifications mainly occur to normal columnar and severe squamous cells, given the probability threshold at 0.5.}

\aaa{Finally, for the 7-class problem, an overall error (OE) of 1.6\% is achieved on average. Such an error is lower than errors of previous methods, such as 7.9\% \cite{jantzen2005pap} and 3.9\% \cite{marinakis2009pap}.}

\begin{table*}[!htbp]
\centering
\caption{Performance comparison of our method with previous methods on the Herlev dataset. PSO-1nn: particle swarm optimization for feature selection and 1-nearest neighbor as the classifier. GEN-1nn: genetic algorithm for feature selection and 1-nearest neighbor as the classifier. ANN: artificial neural networks. K-PCA: kernel principal component analysis for dimensional reduction. \zzz{Ensemble: majority voting of three classifiers.} ENS: ensemble classifiers based on Local Binary Pattern (LBP) with different configurations. $dis(S+M)$: discriminative LBP with concatenated sign and magnitude components. In the $H$-$mean$ (\%) column, numbers in () indicate that no such result is present in the literature, so approximate results are calculated based on the corresponding $Sens$ and $Spec$ to enable comparison. \zzz{$^{*}$: The method in \cite{chankong2014automatic} uses leave-one-out cross validation (LOOCV) and excludes the columnar cells, which is not directly comparable to 10-fold (Refs. \cite{jantzen2005pap,plissiti2012importance}) or 5-fold CV (Refs. \cite{marinakis2008particle,marinakis2009pap,nanni2010local,guo2012discriminative,bora2017automated} and our proposed method) that involve all types of cells. $^{\Delta d}$: randomly translate the ground truth nucleus center up to $d$ pixels in both $x$ and $y$ directions. Bold indicates the highest value in each column.}
}
\label{performance}
\begin{tabular}{|p{3cm}|p{1.2cm}|p{1.5cm}|p{1.5cm}|p{1.5cm}|p{1.6cm}|p{1.5cm}|p{1.2cm}|}
\hline
Methods & $k$-fold CV & $Sens$ (\%) & $Spec$ (\%) & $Acc$ (\%) & $H$-$mean$ (\%) & \zzz{$F$-$measure$} & $AUC$\\
\hline
Benchmark \cite{jantzen2005pap} & 10 & 98.8$\pm$1.3 & 79.3$\pm$6.3 & 93.6$\pm$1.9 & (88.0$\pm$NA ) & \zzz{-} & - \\
\hline
PSO-1nn \cite{marinakis2008particle} & 5 & 98.4$\pm$NA & 92.2$\pm$NA & 96.7$\pm$NA & (95.2$\pm$NA ) & \zzz{-} & - \\
\hline
GEN-1nn \cite{marinakis2009pap} & 5 & 98.5$\pm$NA & 92.1$\pm$NA & 96.8$\pm$NA & (95.2$\pm$NA ) & \zzz{-} & - \\
\hline
ANN \cite{chankong2014automatic}\zzz{$^{*}$} & LOO & 99.9$\pm$NA & 96.5$\pm$NA & 99.3$\pm$NA & (98.2$\pm$NA ) & \zzz{-} & - \\
\hline
K-PCA + SVM \cite{plissiti2012importance} & 10 & - & - & - & 96.9$\pm$NA  & \zzz{-} & - \\
\hline
\zzz{Ensemble \cite{bora2017automated}} & \zzz{5} & \zzz{\textbf{99.0}$\pm$NA} & \zzz{89.7$\pm$NA} & \zzz{96.5$\pm$NA} & \zzz{-} & \zzz{93.1$\pm$NA} & \zzz{-} \\
\hline
ENS \cite{nanni2010local} & 5 & - & - & - & - & \zzz{-} & 0.884 \\
\hline
$dis(S+M)$ \cite{guo2012discriminative} & 5 & - & - & - & - & \zzz{-} & 0.964 \\
\hline
\textbf{ConvNet-T} & 5 & 98.2$\pm$1.2 & \textbf{98.3}$\pm$0.9 & \textbf{98.3}$\pm$0.7 & \textbf{98.3}$\pm$0.3 & \zzz{\textbf{98.8}$\pm$0.5} & \textbf{0.99\zzz{8}} \\
\hline
\zzz{ConvNet-T$^{\Delta5}$} & \zzz{5} & \zzz{98.4$\pm$1.3} & \zzz{97.1$\pm$2.3} & \zzz{98.0$\pm$0.8} & \zzz{97.7$\pm$1.0} & \zzz{98.7$\pm$0.6} & \zzz{0.994} \\
\hline
\zzz{ConvNet-T$^{\Delta10}$} & \zzz{5} & \zzz{98.2$\pm$1.2} & \zzz{96.7$\pm$3.1} & \zzz{97.8$\pm$0.5} & \zzz{97.4$\pm$1.2} & \zzz{98.5$\pm$0.4} & \zzz{0.994} \\
\hline
\end{tabular}
\end{table*}

\begin{table}[!t]
\caption{Numbers (and percentages) of correct classification (normal vs. abnormal) for each of the seven cell classes in Herlev dataset.
}
\label{confusionmatrix}
\begin{tabular}{|p{5.4cm}|p{2.4cm}|}
\hline
Cell type & Correct classification\\
\hline
Superficial squamous epithelial & 74 (100\%)\\
\hline
Intermediate squamous epithelial & 70 (100\%)\\
\hline
Columnar epithelial & 94 (95.9\%)\\
\hline
\hline
Mild squamous non-keratinizing dysplasia & 182 (100\%)\\
\hline
Moderate squamous non-keratinizing dysplasia & 145 (99.3\%)\\
\hline
Severe squamous non-keratinizing dysplasia & 189 (95.9\%)\\
\hline
Squamous cell carcinoma in situ intermediate & 147 (98.0\%)\\
\hline
\end{tabular}
\end{table}

\zzz{\subsection{Quantitative Results on HEMLBC Dataset}}
\zzz{Table \ref{performancehemlbc} compares the classification performance between the proposed deep ConvNet-based method and our previous MLP (multilayer perceptron)-based method \cite{zhang2014automation} on the HEMLBC dataset. Although the dataset used in this paper is a subset slightly smaller than that used in \cite{zhang2014automation}, an obvious trend of performance improvement can be observed.}

\begin{table}[!htbp]
\centering
\caption{\zzz{Performance comparison of our method with a previous method on the HEMLBC dataset. ConvNet-T$^{*}$--- It's worth mention that the 989/989 abnormal/normal cells used in this paper is a subset of 1,126/1,126 abnormal/normal cells used in \cite{zhang2014automation} due to data missing of some abnormal cells.} 
}
\label{performancehemlbc}
\zzz{
\begin{tabular}{|p{1.5cm}|p{1.2cm}|p{1.2cm}|p{1.2cm}|p{1.2cm}|}
\hline
Methods & $k$-fold CV & $Sens$ (\%) & $Spec$ (\%) & $Acc$ (\%)\\
\hline
MLP \cite{zhang2014automation} & 5 & 92.5$\pm$NA & 96.0$\pm$NA & 94.3$\pm$NA \\
\hline
\textbf{ConvNet-T$^{*}$} & 5 & \textbf{98.3}$\pm$\textbf{0.7} & \textbf{99.0}$\pm$\textbf{1.0} & \textbf{98.6}$\pm$\textbf{0.3} \\
\hline
\end{tabular}
}
\end{table}

\zzz{\subsection{Computational Speed}}
The average training time of a ConvNet-T running over up to 30 epochs is about 4 hours. Using the $N_{v} \times N_{c}$ = 1000 classification strategy, the testing time for one cervical cell is 3.5 seconds on average.

\section{Discussion}

\subsection{Comparison With Previous Methods}
\label{sub:compare}
The methods in \cite{jantzen2005pap,marinakis2008particle,marinakis2009pap,plissiti2012importance,bora2017automated} follow the traditional cell classification pipeline -- with features derived from manually segmented cytoplasms/nuclei. The techniques presented in  \cite{nanni2010local,guo2012discriminative} perform direct texture classification of the input image. In contrast, our method automatically learns from the input image patch, and thus is not limited by the shortcomings of cell segmentation or feature design. The $Sens$ values of previous methods are slightly higher than those from our method (\zzz{99.0}\% vs. 98.2\% under 5-fold CV). Such high $Sens$ results mainly from the imbalanced data distribution -- number of abnormal cells $\sim 3$ X higher than the number of normal cells -- which induces the classifier to predict more cells as abnormal. \zzz{High $Sens$ even at the expense of fairly low $Spec$ is required for specimen level diagnosis, as all positives will be reexamined by human experts.}
However, considering the abundance of normal cells (up to 300,000) in a Pap smear slide, the resulting lower $Spec$ will generate many false positives in clinical practice. For example, a 92\% specificity \cite{marinakis2008particle,marinakis2009pap} will result in about 24,000 false positive cells. As a result, extensive and tedious targeted reading from a human observer will be necessary to refine the accuracy of the screening. Our approach substantially decreases the number of false positives. Although there are still about 1.7\% false positives, they only come from columnar epithelial cells (Table \ref{confusionmatrix}). \zzz{Actually, the differentiation between some columnar epithelial cells and (severe) abnormal cells are also challenging for experienced pathologists.} Our method perfectly eliminate\zzz{s} the majority types of normal cells (superficial and intermediate epithelial) in a specimen, and thus alleviates the labor burden of targeted reading and potentially reduces screening errors. \zzz{Compared to our previous MLP method \cite{zhang2014automation} on HEMLBC dataset, the deep ConvNet method achieves both higher $Sens$ and $Spec$ at cell level. Actually, the automation-assisted screening system \cite{zhang2014automation} built upon the MLP method has a satisfyingly high $Sens$=88\% and a perfect $Spec$=100\% at slide level by pathologist's targeted reading. Therefore, our new method has a high potential to further improve the $Sens$ of screening system while reducing the labor burden of targeted reading.}

\subsection{Advantages of the Proposed Method}
1) The proposed method is designed for robust automated screening applications, since it only requires a coarse nucleus centroid as input (no cytoplasm/nucleus segmentation is needed). Our experiments indicate that the proposed image patch based cell classification is robust to inaccurate detection of nucleus centroid \zzz{(refer to ConvNet-T$^{\Delta 5}$ and $^{\Delta 10}$ in Table \ref{performance})}. Some examples can be seen in Fig. \ref{fig4} and Fig. \ref{fig5}, where most of the image patches are not centered on the exact nucleus centroids, but are still assigned reasonable abnormal scores by our method. 
2) Moreover, our method is able to distinguish the abnormal and normal cells even for some ``difficult" cases. For example, the two columnar epithelial cells in the third column in Fig. \ref{fig5} appear to exhibit a far greater level of abnormality than a severe dysplasia cell (lower one in the third column in Fig. \ref{fig4}), as these columnar cells have much larger nuclei or nonuniform chromatin distributions. Unlike traditional morphology/texture based classification methods, which simply classify both cells as abnormal, our method provides a much higher abnormal score (0.89) for the severe dysplasia cell than the two columnar cells (0.01 and 0.21). This indicates that the ConvNet-T captures some latent but essential features embedded in the cell images. 
3) Finally, the deep learning based method has high $Sens$ and \zzz{especially high} $Spec$, and produces the highest \zzz{performances} on \zzz{a Pap-stained Pap smear (Herlev) and a H\&E stained liquid-based cytology (HEMLBC) datasets}. Such a strong performance has the potential to boost the development of automation-assisted reading systems in primary cervical cancer screening. 

\subsection{Limitations}
Despite its high performance, our method demonstrates a few limitations hindering its inclusion in existing cervical screening systems. 
1) Classification of a single patch requires 3.5 seconds, which is far too slow in a clinical setting. One could address this issue by eliminating the image patch augmentation step (100 variants per patch) from the testing phase, thus reducing speed to 0.035 seconds while compromising accuracy by only $\sim 1$\%. 
2) Despite high classification \zzz{accuracy} on the Herlev dataset, our method \zzz{misclassify a few severe dysplasia (4.1\%) and carcinoma (2\%) cells as normal (Table \ref{confusionmatrix} and Fig. \ref{figdistrib})}. As shown in Fig. \ref{fig6}\zzz{(a)}, \zzz{two dark stained carcinoma nuclei and} a very large severe \zzz{dysplasia} nucleus \zzz{are} incorrectly classified as normal. \zzz{An ideal screening system should not miss such severe abnormalities.} To \zzz{better handle} such mis-classifications, cytoplasm/nucleus segmentation based features could be integrated into the system, either via deep learning or via ``TBS" rules. \zzz{Furthermore, both Herlev and HEMLBC are mainly consisted of expert-selected ``typical" cells. The real life situation is more complex so more investigations are needed before transferring the results of this study to practice. For example, refer to the last two false positive cells in Fig. \ref{fig6}(b), they are from NILM slides, but it's hard to tell whether: the first one is an abnormal or normal cell due to poor imaging quality, and the second one is a normal atrophic cell or an abnormal cell.} 
3) \zzz{A nucleus center is pre-required for applying our method, and is obtained from the ground truth segmentation in this paper.} However, screening of abnormal cells within a given field-of-view requires \zzz{automated} detection of nucleus centers. Our ongoing study shows that this may easily be achieved using the fully convolutional networks (FCN) \cite{long2015fully,zhang2017combining} for semantic segmentation of cervical cells. \zzz{And we already show that our method is robust to certain amount of inaccurate nucleus center detection.}
4) The current experiments are conducted on \zzz{a majority of} images \zzz{with} individual cells. The effect of overlapping nuclei\zzz{, cell clumps and artifacts} on classification accuracy needs to be analyzed more extensively \zzz{ in the future investigation, since a screening system is expected to able to avoid misclassifying such objects as abnormal cells. Task-specific classifiers (mostly likely relying on deep learning) may be needed to handle these problems \cite{birdsong1996automated,bengtsson2014screening,zhang2014automation}.} 

\section{Conclusion}
This paper proposes a convolutional neural network-based method to classify cervical cells. Unlike previous methods, which rely on cytoplasm/nucleus segmentation and hand-crafted features, our method automatically extracts \zzz{deep} features embedded in the cell image patch for classification. It consists in extracting image patches coarsely centered on the nucleus as network input, transferring features from another pre-trained model into a new ConvNet for fine-tuning on the cell image patches, and aggregating multiple predictions to form the final network output. The proposed method yields the highest performance on \zzz{both} the Herlev Pap smear \zzz{and the HEMLBC liquid-based cytology datasets}, compared to previous methods. We anticipate that a segmentation-free, highly accurate cervical cell classification system of this type is promising for the development of automation-assisted reading systems for primary cervical screening.

\section*{Acknowledgments}
This work was supported in part by the Intramural Research Program at the NIH Clinical Center, and the National Natural Science Foundation of China (81501545). The authors thank Nvidia for the TITAN Z GPU donation.

\ifCLASSOPTIONcaptionsoff
  \newpage
\fi

\bibliography{ref_cervical}   
\bibliographystyle{IEEEtran}   

\end{document}